\documentclass[9pt,conference,english]{IEEEtran}
\usepackage[T1]{fontenc}
\usepackage[utf8x]{inputenc}
\usepackage{graphicx}
\usepackage{subfig}
\usepackage{babel}
\usepackage{ amssymb }
\usepackage{amstext}
\usepackage{amsmath}
\usepackage{ bbold }
\usepackage{accents}
\newcommand{\ubar}[1]{\underaccent{\bar}{#1}}
\newcommand{\opf}{}

\begin{document}

\title{Supervised Learning for Optimal Power Flow as a Real-Time Proxy}

\author{\IEEEauthorblockN{Rapha\"{e}l Canyasse,
		Gal Dalal, Shie Mannor}
	\IEEEauthorblockA{Department of Electrical Engineering \\
		Technion, Israel Institute of Technology,
		\{raphael.can,gald\}@tx.technion.ac.il, shie@ee.technion.ac.il}
	}

\maketitle
	\begin{abstract}

		In this work we design and compare different supervised learning algorithms to compute the cost of Alternating Current Optimal Power Flow (ACOPF). The motivation for quick calculation of OPF cost outcomes stems from the growing need of algorithmic-based long-term and medium-term planning methodologies in power networks. Integrated in a multiple time-horizon coordination framework, we refer to this approximation module as a \emph{proxy} for predicting short-term decision outcomes without the need of actual simulation and optimization of them.  Our method enables fast approximate calculation of OPF cost with less than 1\% error on average, achieved in run-times that are several orders of magnitude lower than of exact computation. Several test-cases such as  IEEE-RTS$96$ are used to demonstrate the efficiency of our approach.

	\end{abstract}

\section{Introduction}
Alternating Current Optimal Power Flow (ACOPF) is solved hourly or intra-hourly by system operators (SO) world-wide as part of real-time operation process, to ensure safe and robust operation. 
The resulting mathematical problem is computationally complex, as it is a non-linear, non-convex optimization program.

In recent years, there is a growing interest in devising smart algorithmic solutions for long-term optimal planning in different tasks; two examples are system development and asset management \cite{GARPUR_2.2,dalal2016distributed}. Such planning tasks are becoming increasingly complicated as intermittent generation capacity increases \cite{GWEC}. As a result, more stochasticity is involved in power system operation to the extent where sophisticated prediction and optimization methods are crucial \cite{GARPUR_2.1}. The complex coordination between long-term operation and short-term control, combined the high uncertainty in long time-horizons  makes long-term planning extremely challenging. As demonstrated in \cite{dalal2016distributed}, solving an extensive amount of OPF problems to mimic short-term decision-making does not scale well to realistic grids, with thousands of nodes, generators and loads. The computational burden increases in planning for horizons of months to years.

Stochastic optimization is often the tool used for planning under uncertainty. When Monte-Carlo simulation is used, the method involves generation of scenarios in accordance to some probability, which in the case of long-term planning span over months or even dozens of years. In this context, scenario evolution is dependent on the sought plan and contains hourly states of the system and exogenous conditions such as wind generation and consumption. For illustration, consider a maintenance planner assessing several alternatives for next year's proposed outage schedule. To evaluate each of the schedules, a long trajectory of system states needs to be generated, examining the network's ability to comply to a secured OPF in each hour of the trajectory, given the proposed outages of each schedule. He will therefore reproduce different possible network conditions during each of the year's months, in terms of likely nodal wind generation and demand during that month. For each of the reproduced conditions, hourly OPF problems will be solved given the specific future topology of the grid under the outages planned for this month. The planner will conduct this using Monte-Carlo simulation, iterating many times for each of the year's months, per each of the optional outage schedules. The more accurate he wishes the result to be, the more samples of wind and demand values should be drawn from the assumed probabilistic distribution, fed to corresponding secured OPF programs being solved. Each such OPF solution can be used to evaluate the outage schedule in one or more possible ways: counting how many of the OPF programs resulted in a feasible solution; averaging the resulting OPF costs. 

Motivated by the above use-case, in this work we consider the need to quickly solve numerous OPF problem instances, for which the solution accuracy is not of the first priority.

\subsection{Contribution}

Our main contribution is in studying the merits of approximation techniques for OPF feasibility and cost. Other works often assume prior knowledge of the cost function and focus on predicting nodal generation values. Here we solely care about the cost function, thus we are able to perform better in terms of accuracy and run-time gain. We deploy our ACOPF approximation methods on several test-cases, such as IEEE RTS-$96$. For that, we compute OPF feasibility and cost using supervised learning, i.e., an algorithm is given pairs of data samples and their desired output, and generates a functional relation between the two. 

Moreover, we reveal segmentation in regression accuracy that captures spatial clustering of network states; we attribute this behavior to the multimodality originating from different congestion modes. We thus allow for deriving decision rules on if and when to predict OPF costs, or alternatively perform accurate computations. By studying the pros and cons of different known regression techniques,  we allow operators to choose the most appropriate one, i.e, the algorithm that offers the best trade-off between computation time and accuracy.


\subsection{Related work}
Machine learning methods, especially neural networks algorithms, have been used to tackle different problems in power systems as is summarized by Bourguet et al. \cite{bourguet1994artificial}. In the context of Optimal Power Flow (OPF), common approaches rely on supervised learning, where the input is the state of the power system and the output is the solution of the OPF given by any other expert system.

Dated work by Chen et al. \cite{chen1993neural} uses neural networks as a tool for predicting economic dispatch results, tested on a 21 thermal unit system. A more recent work \cite{nakawiro2009combined} trains a neural network for the task of voltage-secured OPF prediction. Additional efforts on tackling economic dispatch and OPF prediction can be found in \cite{siano2012real,imen2015optimal,abdellahpower}. Common to all attempts listed is the multi-dimensional target output of generation values of each generator. This task turns highly challenging as the number of passive network components and generators grow. As opposed to that, we choose to focus on predicting the scalar results of OPF feasibility and cost.

In the context of designing \emph{proxies} for short-term operation as an inner mechanism for long-term planning, recent attempts are reported in \cite{dalal2016hierarchical,duchesne2016machine,dalal2016unit}. These construct proxies for scheduling problems, rather than single-stage decisions. In \cite{dalal2016hierarchical}, reliability is defined based on the ability to recover from any single transmission contingency and approximated using a reinforcement learning algorithm; reliability is again assessed in \cite{duchesne2016machine}, in a day-ahead setting and comparison of supervised learning methods is conducted. The authors of \cite{dalal2016unit} approximate cost values of the scheduling problem using nearest-neighbor classifier, based on network topology as well as wind and load forecasts. These attempts, as well as the one described in this paper, introduce the notion of proxies that approximate short-term decision processes outcomes in a hierarchical multiple time-horizon coordination framework; thus they alleviate the need of actual simulation and optimization of such short-term processes.

\section{OPF Cost and Feasibility Prediction}

Our goal in this work is to find a functional representation to the OPF cost, i.e., the OPF objective function's value at its solution. The ACOPF formulation, as brought in \cite{zimmerman2011matpower} is the following quadratic program, with linear constraints and quadratic cost function:

\begin{flalign} \label{eq:optimals}
&C_{\opf}(x^*,l)=\min_x\frac{1}{2}x^{\intercal}Gx+a^{\intercal}x &&\\
&\text{subject to} \nonumber\\
&\quad\quad g(x,l) = 0 \label{eq:power_balance} \\ 
&\quad\quad h_{\text{min}} \leq h(x,l) \leq h_{\text{max}} \label{eq:operative_limits}
\end{flalign}  

where $x$ denotes active and reactive generation values, along with voltage magnitudes and angles on buses;  $l \in {\rm I\!R_+}^{n_b}$ denotes the vector of loads on all $n_b$ buses;  $C_{\opf}(x^*,l)$ denotes overall OPF cost at the solution $x^*$, which for brevity we also denote by $C^*(l)$; Equation set (\ref{eq:power_balance}) are the nodal power balance constraints; Equation set (\ref{eq:operative_limits}) are the operative limits constraints.

Problem (\ref{eq:optimals}) is feasible when a solution $x^*$ exists given a load vector $l$. Feasibility is denoted by $\delta_{\opf}(l)$; formally,
\[
\delta_{\opf}\left(l\right)=\left\{ \begin{array}{cc}
1 & \text{if a feasible OPF solution exists }\\
0 & \text{otherwise.}
\end{array}\right.
\]

In the rest of the section we introduce the proposed supervised learning approach for predicting $C^*(l)$ and $\delta(l)$, based on load values $l$.

\subsection{Supervised learning}

Supervised learning is a sub-field of machine learning, where prediction of an output value is conducted based on corresponding input values \cite{bishop2006pattern}. Available in supervised learning context is a learning set $\{(l^{(j)},(\delta(l^{(j)}) ,C^*(l^{(j)}))\}_{j=1}^{n}$, composed of $n$ input-output pairs. When the output value is discrete and belongs to a relatively small set, e.g., OPF feasibility $\delta_{\opf}(\cdot)$, the task is referred to and treated as a classification problem. Alternatively, in the case of continuous possible output values, e.g., OPF cost $C^*_{\opf}(\cdot)$,  it is then a regression problem. Both problems are treated in this work. We now refer to the data-set generation process and present the different algorithms used for solving these two problems.

\subsection{Training set generation} \label{sec:training_set_gen}
Supervised learning requires generating dataset of labeled samples, i.e, network states and their corresponding OPF feasibility indicators and their costs.
A network state is defined as the vector of load values on each load bus, i.e., the vector $l =\left(l_{1},\ldots l_{n_b}\right)$. The set of networks states $\{l^{(j)}\}_{j=1}^n$ is generated by uniformly drawing elements from the convex polytope 
\begin{equation}
\left\{ l|l_i\leq\bar{\alpha}\cdot L_i,l_i\geq\ubar{\alpha}\cdot L_i,~ \forall  i=1,\dots,n_b\right\} \label{eq:polytope},
\end{equation} 
 where $\left(\bar{L}_{1},\ldots \bar{L}_{n_b}\right)$ denotes the vector of loads corresponding to the nominal mean demand, and $\bar{\alpha},~\ubar{\alpha}$ are parameters corresponding to minimal and maximal multipliers of these values. We set $\bar{\alpha} = 2$ and $\ubar{\alpha} =0.2$, used for considering both regular and extreme load scenarios. For sampling, we use the Hit\&Run Monte-Carlo algorithm \cite{schmeiser1991hit}.
Once obtaining the sample set, we solve the resulting OPF per each $j$-th sample $l^{(j)}$ and pair it with its feasibility indicator and cost $(\delta(l^{(j)}) ,C^*(l^{(j)})$.






	\subsection{Algorithms }
We now list the algorithms used for classification of feasibility and regression of cost, along with a quick introduction to each.

	\subsubsection{Classification algorithms \label{section:class_algs}}
	The first, simple choice of classification algorithm is the ''trivial classifier'', a predictor that always outputs $1$, meaning assumes the problem is feasible. We also train a naive Bayes classifier with Gaussian kernel \cite{john1995estimating}, that learns a generative model on the data and maximizes posterior distribution for choosing an output; logistic regression \cite{hosmer2004applied} that evaluates the probability of a given output using a logarithm of a linear regression model; decision tree \cite{breiman1984classification} that separates the data to pairs of sets using a feature-value threshold combination; random forest \cite{breiman2001random} and extremely randomized tree \cite{geurts2006extremely} that constitute ensamble variants of decision trees;  neural network \cite{joya2002hopfield} with one hidden fully connected layer of $10$ neurons that well captures non-linear function representations.

	\subsubsection{Regression algorithms}

	A basic regression algorithm we consider first is linear regression \cite{kutner2004applied}, that has linear prediction complexity in the number of features. In addition, we also examine a multi-model extension of linear regression -- piece-wise linear regression \cite{vieth1989fitting} which computes different linear regression models on subsets of the dataset; Gaussian process regression \cite{Rasmussen06gaussianprocesses} with matern32 and ardmatern32 kernels for its capability of estimating any continuous function arbitrarily well due to the Gaussian priors which assign mass to every neighborhood of points in the training set, at the price of prediction time that is dependent on training set size; neural network \cite{joya2002hopfield} with one hidden fully connected layer of $10$ neurons.

\section{Experiments} 

We first discuss the metrics used in our evaluation methodology to compare the different utilized methods and then present our experimental results.

\subsection{Evaluation methodology}

The goal of a supervised learning algorithm is to generalize well on samples that have not been presented to it during training.  We thus randomly divide our labeled dataset into a training set which consists of $80\%$ of the dataset, and a test set which consists of the remaining $20\%$.

Evaluating performance of an algorithm depends on its role; classification algorithms are evaluated using the accuracy metric whereas regression algorithms are evaluated using average relative error. We denote by $\delta^{*}_{\opf}(l)$ and  $C^{*}_{\opf}(l)$ the true feasibility and  optimal cost values as calculated by an OPF solver, and by $\hat{\delta}^{*}_{\opf}(l)$ and $\hat{C}^{*}_{\opf}(l)$  the approximate feasibility and cost computed by our procedure. Classification accuracy is therefore defined as  \[\frac{1}{n_t}\sum_{k=1}^{n_t}\mathbb{1}_{[\delta(l^{(k)})=\hat{\delta}(l^{k})]},\] and regression relative error as \[\frac{1}{n_t}\sum_{k=1}^{n_t} \frac{|\hat{C}^*(l^{(k)}) - C^*(l^{(k)})|}{C^*(l^{(k)})},\] where $n_t$ is the number of test samples.

Furthermore, we compare run-times of predicted versus exact OPF computation. The run-time gain is the ratio between the two.

	\subsection{Experimental results}

	We run our experiments on a MacBook Pro (early 2015) with a $2.7$GHz dual-core Intel Core i5 processor, with $8$GB of memory. 
	All code is written in Matlab \cite{matlab}. We use  Matpower \cite{zimmerman2011matpower}  for network test-cases and OPF calculations.

	In our simulations we consider three test cases: a toy 5-bus case \cite{li2010small}, IEEE RTS-$79$ \cite{subcommittee1979ieee} and IEEE RTS-$96$ \cite{grigg1999ieee}. Peak loads and daily demand profile are based on real data, taken from \cite{UW_website}. The data are then generated as explained in Section \ref{sec:training_set_gen}, sampling from the set described in Eq. (\ref{eq:polytope}). In most figures, the results on the 5-bus case were omitted due to constant high performance, however we still consider it for visualizing the effect explained in Fig. \ref{fig:clustering_case5}.
	
	Training and test set sizes of solved OPFs can be found in Table \ref{tab:set_sizes}. To determine the required dataset sizes we increase it in steps, until the relative error of linear regression stabilizes. Linear regression is chosen due to its fast training and prediction times. We then conduct the following experiments.

		\begin{table}
	\centering
	\begin{tabular}{|c||c|c|}
		\hline

		Case & Training set size  & Test set size \\
		\hline
		\hline 
		Case $5$ & $16000$ & $4000$\\    \hline
		Case IEEE RTS-$79$  & $40000$ & $10000$\\    \hline
		Case IEEE RTS-$96$  & $148000$ & $37000$ \\    \hline
	\end{tabular}
	\caption{Training and test set sizes, used both for cost regression and feasibility classification; these are chosen when reaching a plateau in prediction performance. }
    \label{tab:set_sizes}  
\end{table}

\subsubsection{Feasibility classification}

Each algorithm listed in Section \ref{section:class_algs} is trained on the training set and tested for its accuracy using the test set. Table \ref{tab:Classification_results} summarizes the results. In the case of IEEE RTS-$79$, accuracy of above $98\%$ is obtained using logistic regression, random forest, extra randomized trees and the neural network. Scaling to a larger test-case, IEEE RTS-$96$, exhibits that the only classifier that retains very high performance is the neural network. We attribute this property to neural networks' well-established ability of learning non-linear representations using relatively small data-sets \cite{lanouette1999process}.

	\begin{table*}[h!]
	\centering
	\begin{tabular}{|c||c|c|c|c|c|c|c|}
		\hline

		Test-case & Trivial  & Naive Bayes & Logistic & Decision Tree & Random Forest & Extra-Tree  & Neural Network \\
		\hline
		\hline 
		IEEE RTS-$79$   & $0.8 \pm 0$ & $0.95  \pm 3.5\cdot 10^{-2}$ & $0.99 \pm 2.1\cdot10^{-2}$ & $0.98 \pm 6\cdot 10^{-3}$ & $0.98 \pm 1.7\cdot10^{-2}$ & $0.98 \pm  3.5\cdot10^{-2}$ & $0.99  \pm 4.2\cdot 10^{-2}$ \\    \hline
		IEEE RTS-$96$  & $0.7 \pm 0 $ & $0.73 \pm 4\cdot 10^{-2} $ & $0.84 \pm 8.7\cdot 10^{-2}$ & $0.95 \pm 4.1\cdot 10^{-2}$ & $0.97 \pm 3.6\cdot 10^{-2}$ & $0.97 \pm 7.2\cdot10^{-2}$ & $0.99 \pm 4.7\cdot 10^{-2}$ \\    \hline
	\end{tabular}
	\caption{Feasibility classification accuracy (fraction of correct predictions on the test set) for IEEE RTS-$79$ and IEEE RTS-$96$ }
    \label{tab:Classification_results}
    	\end{table*}

Next, we demonstrate the algorithmic efficiency of training and prediction. Table \ref{tab:Classification_runtime} summarizes actual training run-time of the full training set, and prediction run-time of a single sample on IEEE-RTS$96$ for the different methods inspected. All considered methods' training time is less than $1$ minute, which we consider as negligible when used in the task of predicting thousands of OPF solutions. We witness that for a neural network, which provided the best prediction results, the run-time is an order of magnitude lower than of exact computation, whereas the accuracy sacrifice is extremely low. It is thus encouraging to see that the trade-off between accuracy and run-time is almost inexistent since the best methods are also the quickest.


\begin{table}
	\centering
	\begin{tabular}{|c||c|c|}
	\hline
	
	Method & Training time [s] & Prediction time [s] \\
	\hline
	\hline 
	Gaussian Naive Bayes  & $0.403$ & $5.6\cdot 10^{-6} $   \\    \hline
	Logistic Regression & $ 46.19 $ & $ 2.4\cdot 10^{-6}$             \\    \hline
	Decision Tree & $32.7$ & $2.2\cdot 10^{-6}$            \\    \hline
	Random Forest & $26.1$ & $7.6\cdot 10^{-6}$            \\    \hline
	Extra-Tree & $6.47$    & $9.5\cdot 10^{-6}$                \\    \hline
	Neural Network & $35.3$    & $ 1.3\cdot 10^{-6}$                \\    
	\hline
	\hline
	Exact OPF calculation &  N/A & $0.74$ \\
	\hline
\end{tabular}
	\caption{Average CPU run-time of the full training set, and prediction run-time of a single sample per each classification method, for IEEE RTS-$96$. }
   \label{tab:Classification_runtime}  
\end{table}


\subsubsection{Cost regression}

\begin{figure} 
\centering  
 
\subfloat[Linear Reg.  - IEEE RTS-$79$]{\includegraphics[width=0.15\textwidth]{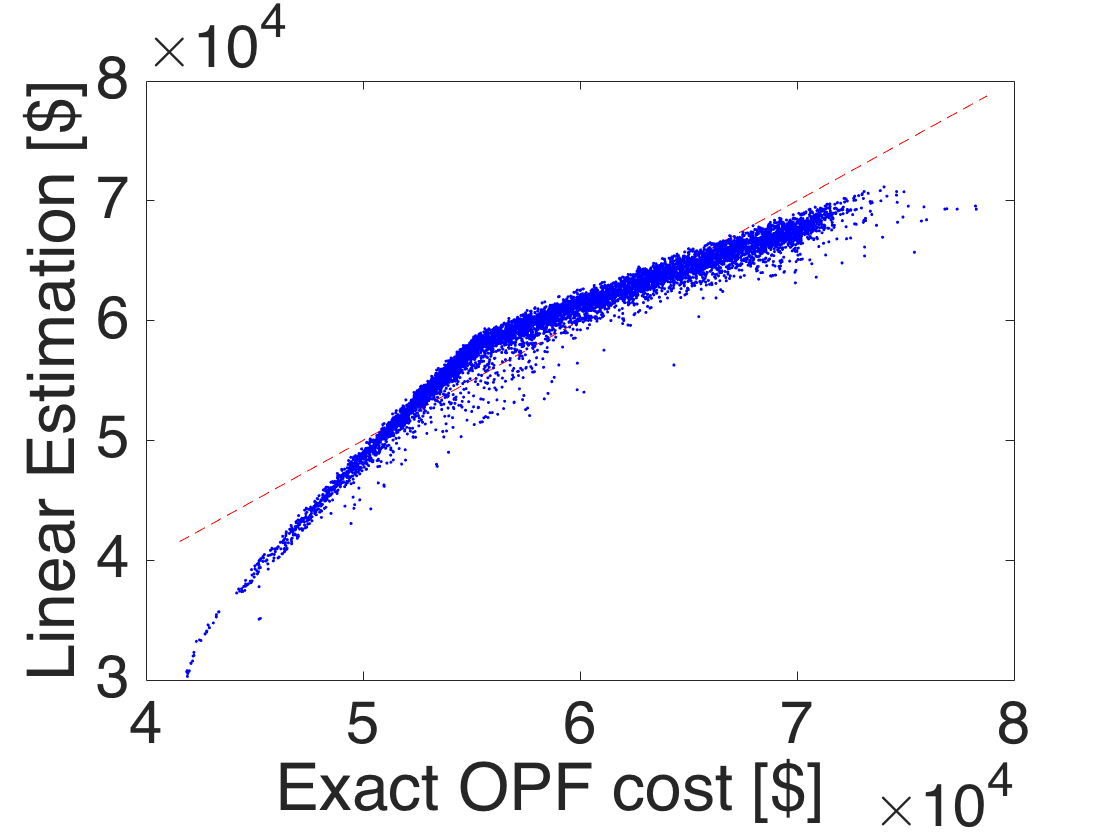}\label{fig:24LinVsExact}}
\hfill
\subfloat[Gauss. Proc. - IEEE RTS-$79$]{\includegraphics[width=0.15\textwidth]{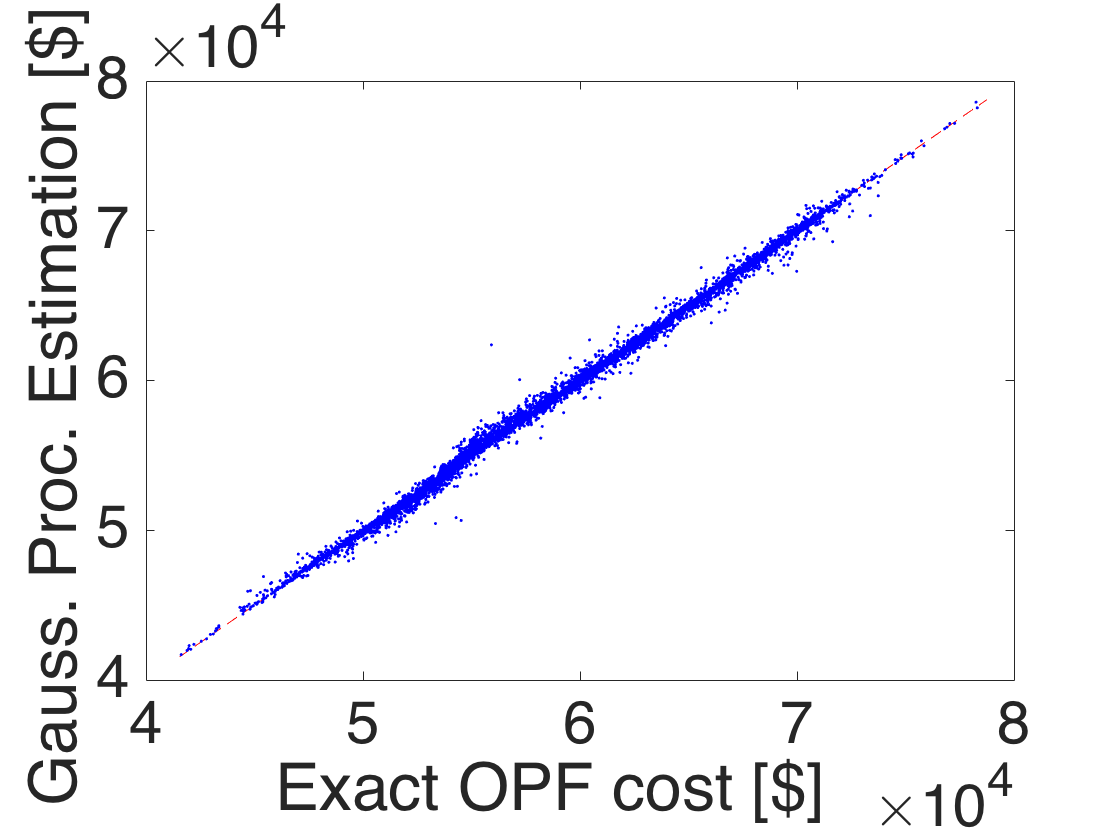}\label{fig:24gprVsExact}}  \hfill
\subfloat[Neural Net  - IEEE RTS-$79$]{\includegraphics[width=0.15\textwidth]{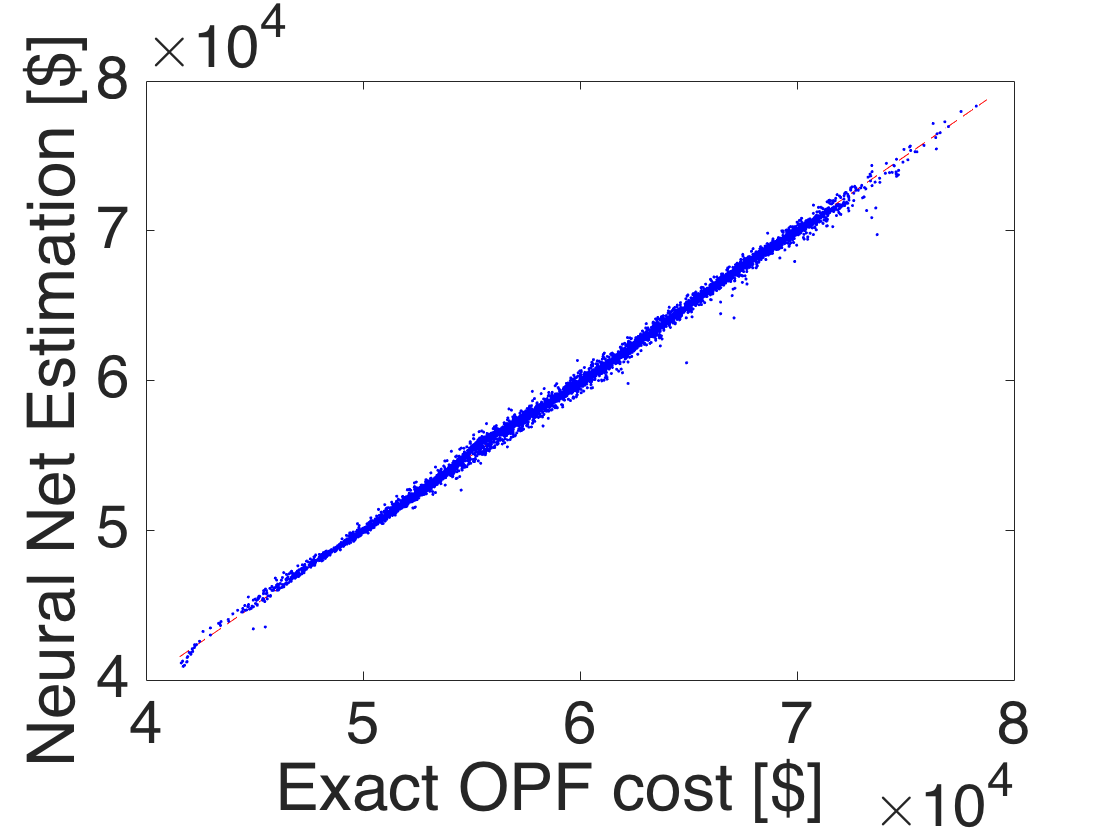}\label{fig:24nnVsExact}}

\subfloat[Linear Reg.  - IEEE RTS-$96$]{\includegraphics[width=0.15\textwidth]{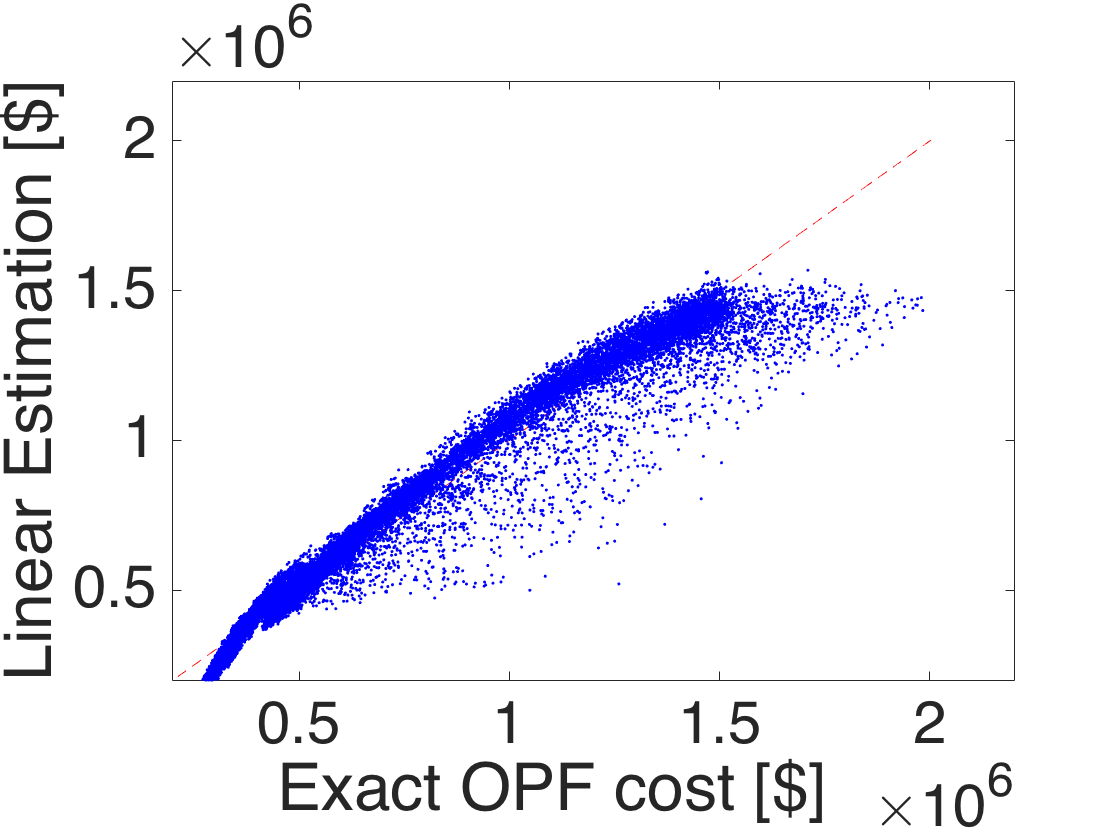}\label{fig:96LinVsExact}}
\hfill
\subfloat[Gauss. Proc. - IEEE RTS-$96$]{\includegraphics[width=0.15\textwidth]{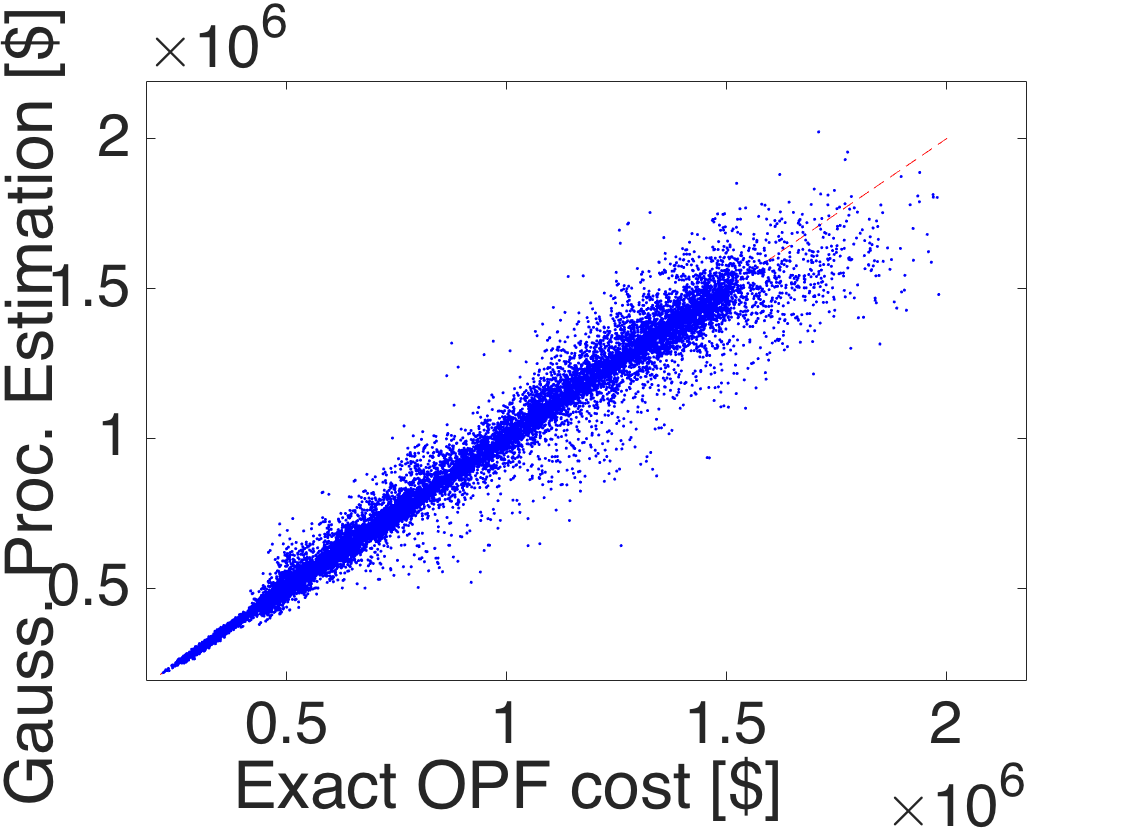}\label{fig:96gprVsExact}}  \hfill
\subfloat[Neural Net  - IEEE RTS-$96$]{\includegraphics[width=0.15\textwidth]{graphics/case24_exactVsNN.png}\label{fig:24nnVsExact}}     
\caption{Scatters of OPF costs, comparing exact solutions to their corresponding predictions, for IEEE RTS-$79$ and IEEE RTS-$96$. \label{fig:PreVsTrue}}

\end{figure}

As in the classification procedure, we begin by training regression algorithms on the training set and then test for their relative error. We first inspect linear regression, for its benefits in simplicity, fast training and prediction time, and explainability. The latter is regarded as the ability to inspect regression coefficients and relate them back to the power system considered. Nonetheless, these desired properties do not justify usage of the algorithm given the poor results shown in Fig. \ref{fig:PreVsTrue}. The figure exhibits scatters of accurate OPF costs versus predicted costs for linear regression, Gaussian process and neural network. 

Careful inspection of linear regression reveals multimodality in the dataset as load and consequently cost increases, that corresponds to different areas of line congestions (we rule out the possibility of piecewise-linear behavior stemming from cost structure, as it is polynomial and not piece-wise in this case). For that reason we also inspected piecewise-linear regression. As will be seen in following paragraphs, it also performed poorly and had long training time; this direction was not further investigated. It can also be seen that Gaussian process and neural network are able to capture the cost function very well, with advantageous results  for neural network in the case of IEEE RTS-$96$.

In addition to inspecting correlation of scatters, we compute the relative error of cost prediction; the results are brought in Table \ref{tab:Regression_Results}. Gaussian Process exhibits the best results for IEEE RTS-$79$, averaging $0.25\%$ relative error. However, it is again demonstrated that a neural network is more appropriate for larger test-cases, as it achieves $0.85\%$ relative error for IEEE-RTS$96$, whereas for Gaussian process it drops to $4.43\%$.

\begin{table*}
	\centering
	\begin{tabular}{|c||c|c|c|c|}
		\hline
		Case & Linear Regression & Piece-wise Regression & Gaussian Process & Neural Network\\    
		\hline
		\hline 
		Case IEEE RTS-$79$  &  $2.85 \% \pm 0.05\% $ & $2.79 \% \pm 0.05\% $& $0.25 \% \pm 0.07\% $  & $0.32 \% \pm 0.07\% $\\    \hline
		Case IEEE RTS-$96$  & $7.6 \% \pm  0.088$  & NA &  $4.43 \%  \pm 0.044 \% $   &  $0.85 \%  \pm 0.013 \% $        \\    \hline
	\end{tabular}
	\caption{Average regression relative error for IEEE RTS-$79$ and IEEE RTS-$96$}
    \label{tab:Regression_Results} 
\end{table*}

	To assess the computational efficiency of our method we compute prediction run-time gain, i.e., the ratio between the algorithm's prediction time and accurate OPF run-time. The results are brought in Table \ref{tab:Regression_run_time_gain}. Piecewise-linear regression for IEEE-RTS$96$ was not conducted due to long training periods. As expected, since linear regression prediction is simply vector multiplication, it achieves extremely large run-time gains -- it is $3.4 \cdot 10^4$ faster than exact computation. Considering an approximation error of $2.85\%$ and $7.6\%$ for the two networks, respectively, this method may turn the desirable one for several applications where such errors are acceptable.  
	
	An additional vital parameter is scalability. As network size increases, linear regression run-time gain increases since accurate OPF solution time grows super-linearly in network components, whereas linear regression grows linearly. It thus outperforms all other algorithms in terms of scalability. In the case of Gaussian process, the method quickly becomes intractable  as network size grows; this is because the training set required grows proportionally  as well, and as explained in Section \ref{section:class_algs},  prediction time of Gaussian process is dependent on the training set size. Neural networks, on the other hand, are scalable as their computational complexity as a function of network components is lower than of OPF calculation, thus showing increased run-time gain for IEEE-RTS$96$ compared to IEEE-RTS$79$.

	\begin{table*}
	\centering
	\begin{tabular}{|c||c|c|c|c|}
		\hline

		Method & Linear Regression & Piece-wise Regression & Gaussian Process & Neural Network \\    
		\hline
		\hline 
		Case IEEE RTS-$79$ & $2.3\cdot10^4$ & $348$ & $13.3$ & $10.8$ \\
		\hline
		Case IEEE RTS-$96$ & $3.4\cdot10^4$ & NA & $2.8$ & $14.3$ \\ 
	\hline
	\end{tabular}
	\caption{Regression run-time gains for IEEE RTS-$79$ and IEEE RTS-$96$ }
    \label{tab:Regression_run_time_gain} 
\end{table*}

\begin{figure} 
	\centering \includegraphics[scale=0.2]{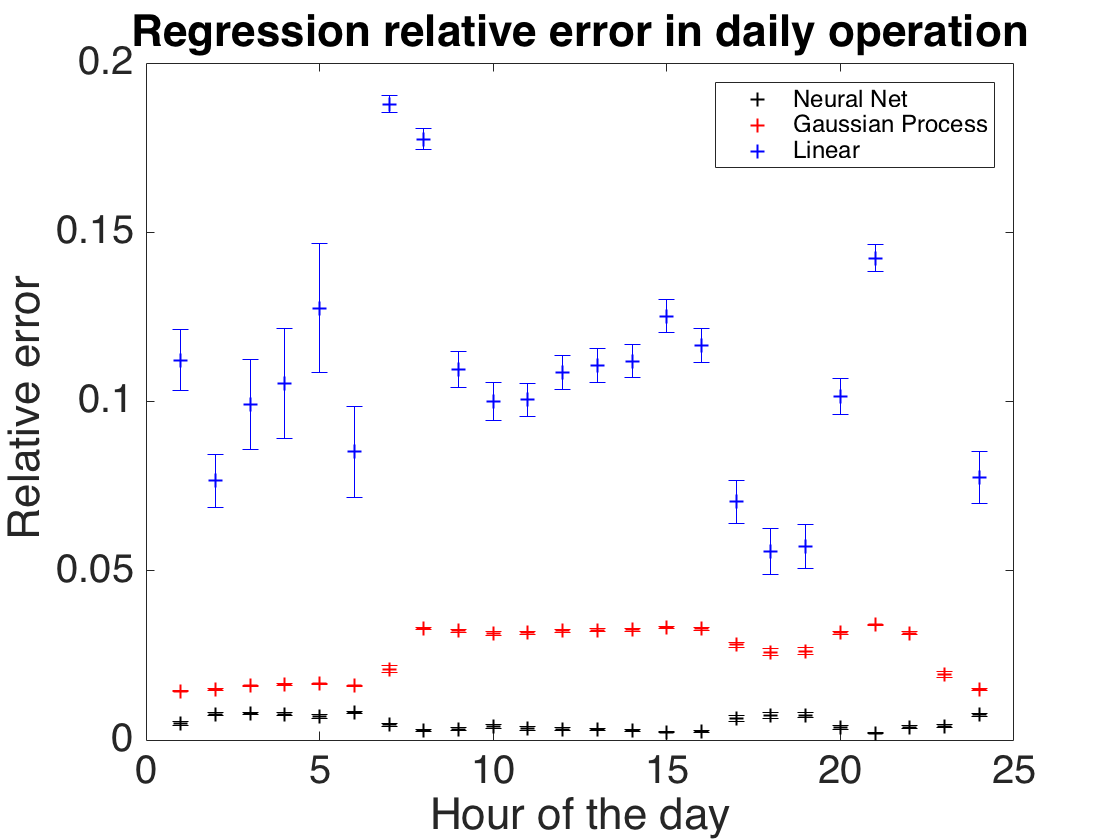}
	\caption{ 	Relative error throughout the evolution of daily load. This is to test the algorithms' sensitivity to changes in loads, without specifically training them on such data.\label{fig:case96_profileComparison}}
	\end{figure}

	In order to inspect the usability of our approach for hourly OPF calculations, such as in the long-term planning assessment scenario described in the introduction, we assess regression accuracy throughout the evolution of daily load. Consequently, we test the algorithms' sensitivity to changes in loads, without specifically training them on such data. The daily load profile is based on historical US data, adopted from \cite{UW_website}; we simply compute the mean profile across multiple trajectories and use it to scale the peak load according to the hour of the day. The results are found in Fig. \ref{fig:case96_profileComparison}. They exhibit high sensitivity of linear regression to new data, performing the worst among the algorithms tested by producing maximal relative error of nearly  $20\%$ at morning time. Gaussian process achieves a maximal error of $3.5\%$, which remains consistently high during the day. Neural network greatly outperform other algorithms, as it does not achieve more of $ 1.2\% $ error. Overall, the two last techniques seem to be well suited for computing the cost in a real world situation, being able to generalize well on new data.

\begin{figure}  
\centering  
\subfloat[Gaussian Process: low, medium and high error intervals are  ${[0\%,0.12\%]},~{[0.12\%,0.38\%]}$,
${[0.38\%,1.4\%]}$.]{\includegraphics[width=0.33\textwidth]{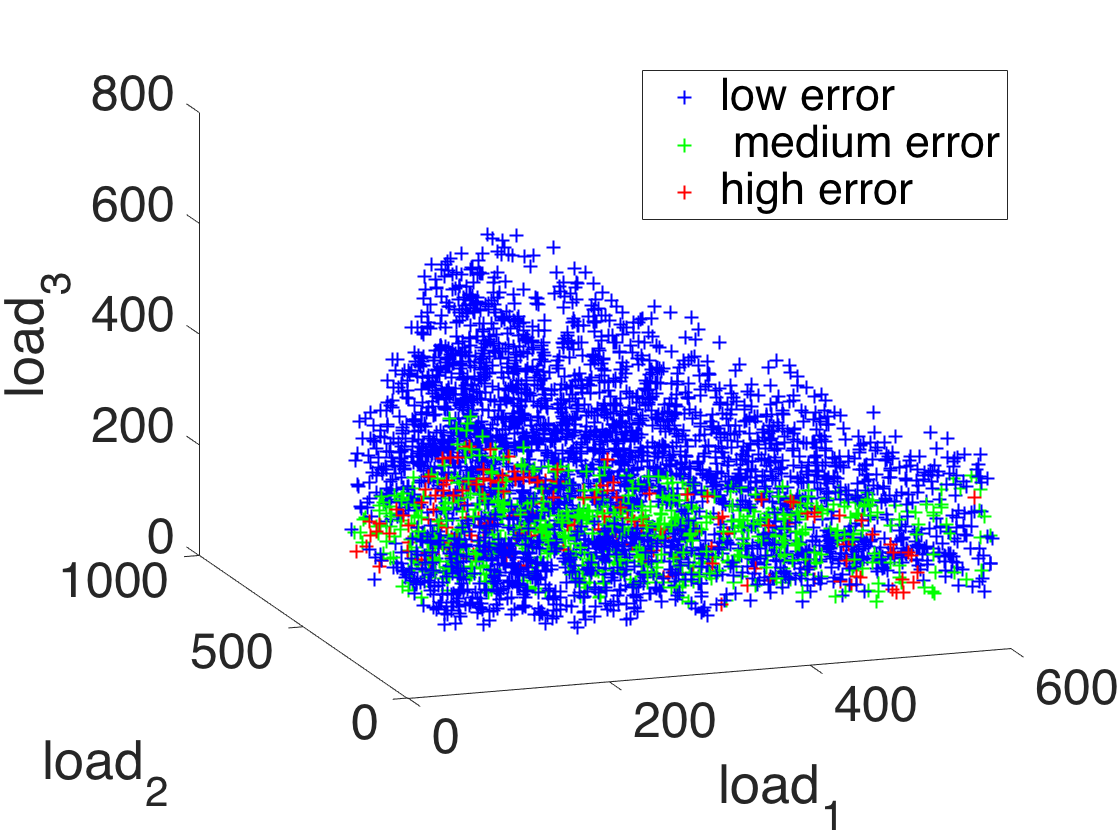} \label{fig:db_gpr_clustering_ac_case5}}   
\hfill   
\subfloat[Linear Regression: low, medium and high error intervals are  ${[0\%,13\%]},~{[13\%,55\%]}$, ${[55\%,196\%]}$. ]{\includegraphics[width=0.33\textwidth]{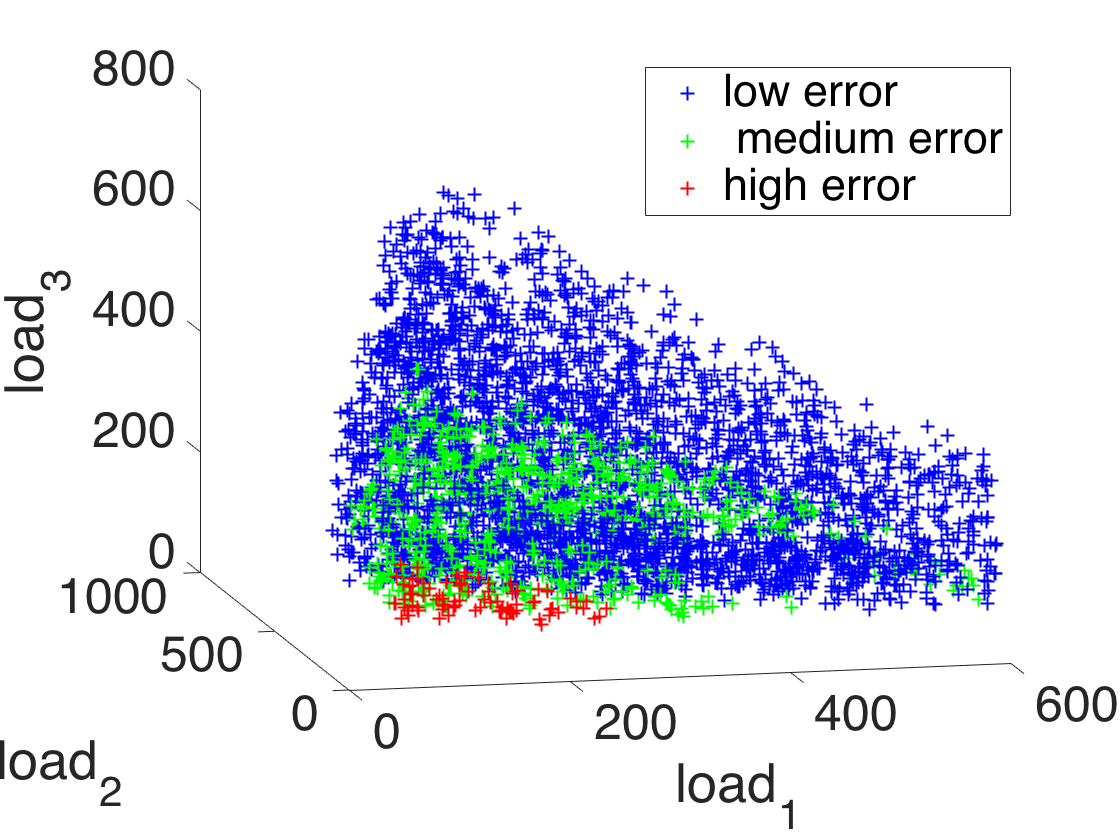}\label{fig:db_lr_clustering_ac_case5}}
\caption{Cost regression approximation quality as a function of load values in the different buses. The segmentation is performed using thresholds that are automatically determined using the $K$-means clustering algorithm. \label{fig:clustering_case5}}
\end{figure}

	Lastly, we investigate approximation quality as a function of load values in the different buses. For that purpose, relative regression error is segmented into three groups of low, medium and high errors. The segmentation is performed using thresholds that are automatically determined using the $K$-means clustering algorithm. Fig. \ref{fig:clustering_case5} contains plots of the results for Gaussian process and linear regression in 3D. The 5-bus test-case is used in that experiment, as it only contains three loads and thus easily visualized. The plots in both algorithms exhibit clear clustering of approximation quality in specific sets of linear load combinations. We again attribute this behavior to the multimodality originating from different congestion modes. Experimenting with larger test-cases required dimensionality reduction; we therefore applied Principal Component Analysis (PCA) on the loads and inspected the resulting 3D plots. We noticed that projecting to the principal component space skewed the segmentation, rendering this visualization method useless in the setting we experimented with.

	\section{Conclusion}
In this work we use supervised machine learning methods to quickly predict two outputs of optimal power flow calculation. We show that for the test-cases inspected, extremely high accuracy for both feasibility classification and cost regression are obtainable.

At times, the accuracy vs. runtime trade-off is not be resolved by solely focusing on the former. The potential overall gain in CPU time is the fundamental advantage of this method, when used in the context of long-term assessment/control applications. We enable predicting the feasibility of OPF problems in run-time that is several orders of magnitude lower than exact calculations, and with accuracy that is occasionally above $98\%$. Thus, we allow for a dramatic saving in CPU time of up to $34$K as shown in Table \ref{tab:Regression_run_time_gain}. Additionally, the method allows for solution cost computation without prior knowledge of the cost function.

The comparison between different regression methods reveals a trade-off in accuracy and run-time gain. Linear regression, for example, demonstrates a run-time gain of roughly $2$K, in both training and prediction, over neural networks. Albeit neural networks are often considered the popular choice, it may not always fulfill the needs of fast enough computation when approximation quality is not of first priority. 

Further research will investigate the usage of deep networks with convolutional-type layers tailored to a power network rather than the usual images use-case. It will also consider an active learning framework, possibly in a semi-supervised setting, where confidence is maintained on prediction accuracy; when confidence high, prediction result should be used. However, when it is below a certain threshold, exact OPF calculation will be made instead.

\bibliographystyle{IEEEtran}
\bibliography{ISGT2017.bib}

\end{document}